\documentclass[letterpaper, 10 pt, conference]{ieeeconf}
\IEEEoverridecommandlockouts                              % This command is only needed if 
\usepackage{graphics} % for pdf, bitmapped graphics files
\usepackage{epsfig} % for postscript graphics files
\usepackage{times} % assumes new font selection scheme installed
\usepackage{amssymb}  % assumes amsmath package installed
\usepackage{subfigure}
\usepackage{caption}
\usepackage{url}
\usepackage{multirow} 
\usepackage{amsmath} 
\usepackage{xcolor}
\usepackage{tabu} 
\usepackage{bm}   
\usepackage{cases}
\usepackage{subfigure}
\usepackage{booktabs}
\usepackage{epstopdf}
\usepackage[colorlinks,linkcolor=blue]{hyperref}
\usepackage{algorithmic}

\newcommand{\be}{\begin{equation}}
\newcommand{\ee}{\end{equation}}
\newcommand{\bea}{\begin{eqnarray}}
\newcommand{\eea}{\end{eqnarray}}
\newcommand{\ba}{\begin{array}}
\newcommand{\ea}{\end{array}}

\newcommand{\re}[1]{(\ref{#1})}

\usepackage[ruled]{algorithm2e}

\title{\LARGE \bf Learning compliant grasping and manipulation by teleoperation \\
with adaptive force control}

\author{Chao Zeng$^{1\dagger}$, Shuang Li$^{1\dagger}$, Yiming Jiang$ ^{2} $, Qiang Li$^{3}$, \\
 Zhaopeng Chen$^{1\ast}$, Chenguang Yang$^{4}$, and Jianwei Zhang$^{1}$
\thanks{This work was supported jointly by  the German Research Foundation (DFG)  and the National Science Foundation of China (NSFC) in project DEXMAN under grant No.410916101, and the project in  Cross Modal Learning  TRR-169.}
\thanks{$^{1}$TAMS group (Technical Aspects of Multimodal Systems), Department of Informatics, Universit¨at Hamburg, Germany.  $^{2}$National Engineering Laboratory for Robot Visual Perception and Control, Hunan University, China. $^{3}$Center for Cognitive Interaction Technology, Bielefeld University,  Germany. $^{4}$Bristol Robotics Laboratory, University of the West of England, UK.} 
\thanks{$^{\dagger}$ The first two authors contributed equally to this work.}
\thanks{ $^{\ast}$ Corresponding author, { Email: zhaopeng.chen@uni-hamburg.de}}
}

\begin{document}

\maketitle

\begin{abstract}
%\boldmath
%In this work, we propose a learning-control based approach for acquiring compliant grasping and manipulation skills of a multi-finger robot hand. This approach takes the depth image of the human hand as input and generates the desired force commands for the robot. The vision-based TeachNet teleoperation system developed in our previous work maps the pose of the human hand to the joint angles of the Shadow hand in real-time, given a depth image of the bare human hand.  To endow the robot hand with compliant human-like behaviours, a novel adaptive force controller is designed to predict the desired force control commands based on the pose difference between the robot hand and the human hand during the demonstration. The force controller is derived from a computational model of the biomimetic control strategy in human motor learning, which allows  adapting  the control variables (impedance and feedforward force) online during the execution of the reference joint angles. The simultaneous adaptation of the impedance and feedforward profiles enables the robot to interact with the environment compliantly. Our approach has been verified in four different types of tasks in physics simulation, i.e., grasping, opening-a-door, turning-a-cap, and touching-a-mouse, and has shown more reliable performances than the existing position control and the fixed-gain-based force control approaches. 

In this work, we focus on improving the robot's dexterous capability by exploiting visual sensing and adaptive force control. TeachNet, a vision-based teleoperation learning framework, is exploited to map human hand postures to a  multi-fingered robot hand. We augment TeachNet, which is originally based on an imprecise kinematic mapping and position-only servoing,  with a biomimetic learning-based compliance control algorithm for dexterous manipulation tasks. This compliance controller takes the mapped robotic joint angles from TeachNet as the desired goal, computes the desired joint torques. It is derived from a computational model of the biomimetic control strategy in human motor learning, which allows adapting the control variables (impedance and feedforward force) online during the execution of the reference joint angle trajectories. The simultaneous adaptation of the impedance and feedforward profiles enables the robot to interact with the environment in a compliant manner. Our approach has been verified in  multiple  tasks in physics simulation, i.e., grasping, opening-a-door, turning-a-cap, and touching-a-mouse, and has shown more reliable performances than the existing position control and the fixed-gain-based force control approaches.

%First, the approach is  built on the vision-based TeachNet teleoperation system developed in our previous work. Given a depth image of the human hand collected by a camera, TeachNet maps the pose of the human hand pose to the joint angles of the Shadow hand in real-time.  Then, a novel adaptive force controller is designed to predict the desired force control commands based on the pose difference between the robot hand and the human hand during the demonstration. The force controller is derived from a computational model of the biomimetic control strategy in human motor learning, which allows  adapting  the control variables (impedance and feedforward force) online during the execution of the reference joint angles. The simultaneous adaptation of the impedance and feedforward profiles enables the robot to interact with the environment in a compliant manner. Our approach has been verified in four different types of tasks in simulation, i.e., gasping, opening-a-door, turning-a-cap, and touching-a-mouse, has shown more reliable performances, compared with the existing  position control and fixed-gain based force control approaches. 
\end{abstract}

%\begin{IEEEkeywords}
%
%Adaptive impedance/force control; Robot control; Impedance learning; Human-robot interaction; Robotics
%\end{IEEEkeywords}

\section{Introduction}

%Learning how to perform a task compliantly and dexterously is a key topic in robot learning. The goal of this topic is to enable a robot to acquire human-like behaviours during the execution of a specific task. Several attempts have been made toward this goal in recent years, but several issues still remain to be addressed urgently, especially when it comes to the grasping and manipulation by a multi-finger robot hand \cite{ficuciello2019vision}. Most of the state-of-the-art works in robotic grasping  mainly focus on object recognition or grasping motion planning (see, e.g., \cite{liang2019pointnetgpd,  mahler2019learning}). They usually control robot hands using a very straightforward way: close the grippers or fingers to grasp the target.  The task dynamics during the grasping process is often  neglected. However, this kind of control strategy may be unsuitable for tasks where compliance is required such as opening-a-cap, due to the lack of adaptation and flexibility. Yet, we humans can spontaneously adapt our hand pose and force to interact with environments in a compliant manner during daily manipulation tasks. Similarly, if we would like to endow a robot with human-like skills,  one promising solution is therefore to develop adaptive control strategies that allow the robot to compliantly deal with physical and dynamical  interactions with  the environment. 

It is a major goal in robotic manipulation research to augment robots with human-like dexterous and compliant behaviour for many tasks in everyday life. In recent years numerous attempts have been published towards this goal, but some issues are still not fully addressed yet, especially for grasping and manipulation with a multi-finger robot hand \cite{ficuciello2019vision, zhuang2019shared, ruppellearning}. Most of the state-of-the-art works in robotic grasping mainly focused on object recognition or grasping motion planning (see, e.g., \cite{liang2019pointnetgpd,  mahler2019learning, lu2020multifingered}), and the robotic hand was controlled in a binary way--closing fingers to grasp the object. The task dynamics during the grasping process were often neglected. This kind of control strategy is not suitable for a complex task in which a fine-tuned grasp posture and compliance fingers motions are needed. Yet, we humans can spontaneously adapt our hand pose and force to interact with environments in a compliant manner during daily manipulation tasks. Consequently, if we would like to endow a robot with human-like skills,  one promising solution is therefore to develop adaptive control strategies that allow the robot to compliantly deal with physical and dynamical  interactions with  the environment. 

%\begin{figure}[t!]
%\centering
%\includegraphics[scale=0.1]{pic/pipeline3.jpg}
%\caption{\small The overview (a) and pipeline (b) of the proposed approach. The human demonstrator guides the robot hand to complete a task through the vision-based teleoperation. A camera is used to track the hand pose during the demonstration.  The TeachNet model is used to map the human hand pose to the robot hand joint angles. Unlike our previous work which sends position commands to the robot hand directly, this work develops a force predictor to generate the desired force commands to control the robot hand in an online manner.}
%\label{pipeline}
%\end{figure}

\begin{figure}[t!]
\centering
\includegraphics[scale=0.5]{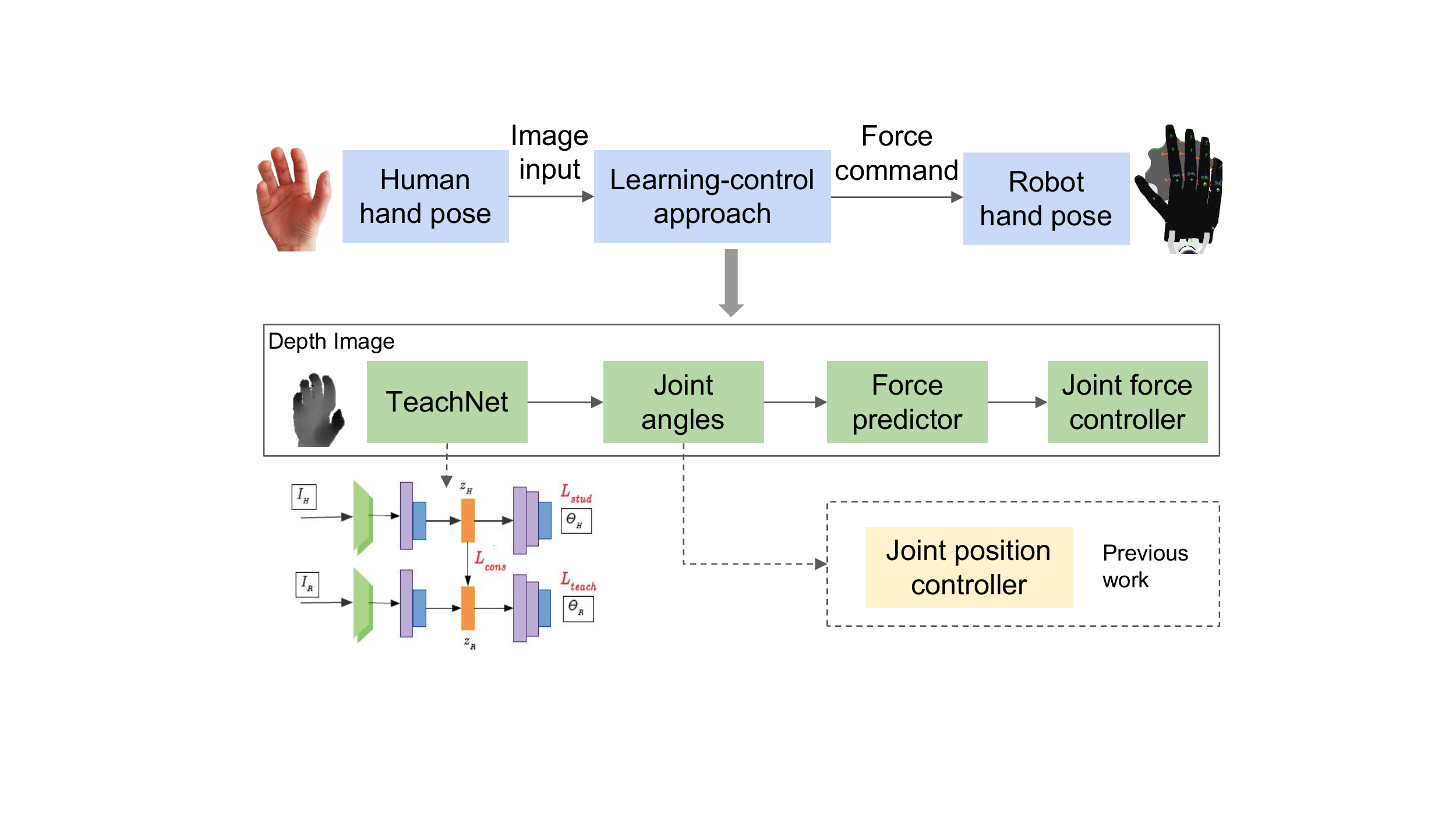}
\caption{\small The pipeline of the proposed learning-control approach. The human demonstrator guides the robot hand to complete a task through the vision-based teleoperation. A camera is used to track the hand pose during the demonstration.  The TeachNet model is used to map the human hand pose to the robot hand joint angles. Unlike our previous work which sends position commands to the robot hand directly, this work develops a force predictor to generate the desired force commands to control the robot hand in an online manner.}
\label{pipeline}
\end{figure}
%Vision-based

Teleoperation is considered as an efficient way for the robot to imitate manipulation behaviours from humans. Recently, markerless vision-based teleoperation offers several advantages for multi-finger robots such as a low cost and no obstructions due to measurement devices. In a typical teleoperation system, the human demonstrator's behaviours are captured through a motion tracking device and then mapped into the robot's motion policies in the Cartesian space or joint space. In this case, the robot is often required to be controlled under the position control mode. As a matter of fact, several studies  have revealed that force control strategies could obtain good performances for robot compliant manipulation (see, e.g., \cite{li2018force, van2018comparative}). Thus,  it is reasonable to integrate an adaptive force control strategy into a vision-based teleoperation system such that we can bring their advantages together for the manipulation of the multi-fingered robotic hand.

In this work, we propose a learning-control approach combining a vision-based teleoperation system with adaptive force control, allowing us to take an image as the input and output  the desired force commands for the robot hand. The pipeline of the proposed approach is shown in Fig.~\ref{pipeline}. For the learning part, we employ the end-to-end model TeachNet developed in the previous work \cite{li2019vision} to learn the mapping relation between the human hand pose and the joint angles of the Shadow robot hand. At runtime, a camera is used to  collect a depth image of the hand, and the TeachNet  estimates the desired robot joint angles based on the image.  To deal with task dynamics during hand grasping and manipulation, we develop an adaptive force control strategy which can predict the next-step desired control force command based on the desired joint angles and the robot current states. Our force controller is derived from the computation model  inspired by the human motor learning principles. The control variables in the controller, i.e., impedance and feedforward terms are simultaneously adapted online and combined  to generate the force/torque commands which are subsequently sent to the robot hand in the joint space.  We need to mention that the proposed controller does not exploit tactile feedback, but is a biomimetic-inspired open-loop compliance method. Even so, the physics simulation experiments show that the grasp stability with proposed approach is better than the existing position control and the fixed-gain-based force control approaches.
%Several comparative simulation tasks are conducted to verify the effectiveness of our approach. 

\section{Related work}
%Previous studies related to our work can be mainly categorised as \textit{learning of robotic complaint movements} and \textit{biomimetic control based on human motor learning}.
%
%\subsection{Learning of robotic complaint movements}
%Learning compliant grasping and manipulation from demonstration with adaptive force control 

\subsection{Markerless vision-based teleoperation }
Vision-based teleoperation systems have been widely used for robots learning skills from human demonstrations \cite{ficuciello2019vision, handa2020dexpilot, li2020mobile}. Typically, a human demonstrator and a robot together constitute a human-in-the-loop leader-follower system, in which a motion-tracking device (such as LeapMotion and  Kinect) is usually utilized to capture the human demonstrator's movements. Then, the demonstrated motion is further mapped to the robot's workspace to enable the imitation of the human behaviours. Compared with wearable device-based  teleoperation techniques (such as a data glove or a marker-based tracking system) \cite{cerulo2017teleoperation, fang20183d, han2018online}, markerless vision-based approaches  allow for  natural and unrestricted demonstrations due to non-invasiveness.%**********

% since the human demonstrator's  movements would not be restricted when manipulating an external teleoperation device \cite{hirschmanner2019virtual}. 

A core issue in this teleoperation system is how to map the human hand pose to the robot hand pose. Since deep learning (DL) techniques offer the advantages of learning  highly non-linear relations, several DL-based hand pose estimation methods have been proposed recently. In \cite{gomez2019accurate, antotsiou2018task}, the authors proposed to track  keypoints of the  human hand, then use retargeting methods (e.g., inverse kinematics) to control the robot. Nevertheless, they usually suffer the time cost of the retargeting post-processing. Thus, our previous work \cite{li2019vision} proposed a neural network model that permits  end-to-end efficient mapping from the 2D depth image representing the human hand pose to the robot hand joint angles. This work aims to extend \cite{li2019vision} to further enable the mapping from the image  to the desired  force commands, which achieve better performances, as observed in the experiments in Section \ref{experiments}.

%To this end,  one way is to estimate the human demonstrator's hand joint angles in a real-time manner, and then to map  them to the follower, i.e., the robot hand, fingers' joints. In this case,  a date glove could be usually used to calculate the human hand joint angles based on the positions of the markers mounted on the glove in the optical tracking frame (see, e.g., \cite{cerulo2017teleoperation, ruppellearning}). An alternative way is to track the positions of human hand finger tips, based on which the robot hand joint angles could be calculated through the hand inverse kinematics. 

%Another way is to track the 

\subsection{Grasping and manipulation based on force/torque control}
An impedance-model based force controller has been used in robotic manipulator control for a  number of physical interaction tasks (see, e.g., \cite{ficuciello2015variable}). However, so far it has not been fully investigated yet to control multifingered robot hands for grasping and manipulation tasks. Recent studies illustrate that  force control strategies increase the grasping stability and robustness \cite{fan2018research, deng2020grasping}, and  achieve a good grasp stability \cite{sommer2016multi} and in-hand manipulation \cite{li2012grasp} with haptic exploration  of multi-finger robotic hand.

In~\cite{wimbock2008analysis}, an object-level impedance controller has been developed and shown the  effectiveness and robustness for robot grasping.  Li \textit{et al.} improved the controller by decomposing the impedance into two parts: one for stable grasping and another for manipulation. Furthermore, the desired impedance is estimated  using supervised learning based on the data collected from the human demonstration in advance \cite{li2014learning}.  Pfanne \textit{et al.} proposed an object-level impedance controller based on in-hand localization,  which improved the ability to avoid contact slippage  through  adjusting the desired grasp configurations \cite{pfanne2020object}. Garate \textit{et al.} proposed to regulate the control of the grasping impedance (stiffness) by regulating both the robot hand  pose and the finger joint stiffness.  By adaptating  the magnitude and the geometry of the grasp stiffness, the desired stiffness profile could be achieved to adapt the hand configuration for stable grasping \cite{garate2018grasp}.

However, these force controllers may not be suitable for our use in a vision-based teleoperation system, where the controller needs to dynamically and quickly respond to the changes of the human hand pose to predict the desired force commands. Consequently, the contribution of this work is to explore the regulation of the impedance (stiffness) and the feedforward term online during the process of robot  grasping or manipulation, which cannot be learned in advance or through exploration.

\subsection{Biomimetic compliant control for robot manipulation} 
Recently, the biomimetic control strategy inspired by the findings of human motor learning in the muscle space has been developed and proved to be an effective way for robot compliant manipulation \cite{zeng2021learning}. It has been discovered in neuroscience that humans can simultaneously adapt the arm impedance and feedforward force  to minimize motion error and interaction force with external environments, under a certain set of constraints \cite{burdet2001central}. Based on this principle, a biomimetic force controller was first proposed in \cite{yang2011human} which allowed the robots to deal with both stable and unstable interactions through the adaptation of the impedance and feedforward term in the force controller. Li \textit{et al.} further improved this controller and implemented it to deal with several physical interaction tasks such as cutting and drilling by a redundant robot manipulator \cite{li2018force}. However, until this work the  biomimetic control strategy has not been utilized for a dexterous robot hand with multiple DOFs. Another contribution of this work is to extend the biomimetic force controller to enable compliant grasping and manipulation from human hand teleoperation.

\section{Methodology}
\label{Methodology}
In this section, we will first briefly introduce how to estimate the robot hand posture (i.e., joints) by teleoperation. Then, we  will elaborate how to generate the desired force/torque control  variables  based on the estimated hand pose.

\subsection{Estimation of robot joint angles  from  vision-based teleoperation}

The goal of this part is to find a proper mapping function $ f_{m} $ from the human hand posture to the joint angles ($ q_{r} $) of the  Shadow robot hand. Here, the human posture at each time step is represented by a 2D image ($ I_{h} $) collected by a camera. Therefore,  we have
\begin{equation}
f_{m}:  I_{h}\in R^{2} \rightarrow q_{r}\in R^{N_{r}},
\label{imageTOjoint}
\end{equation}
where $N_{r}$ denotes the number of DOFs.  In order to determine the function $ f_{m} $, we utilize the neural network architecture named TeachNet to learn the highly nonlinear mapping correlation between $  I_{h}$ and $ q_{r} $. 

%\footnote{Note that $N_{r}$ is  not usually equal to the number of  the joints of a robot hand.}
Here, we briefly summarize the basic utilization procedure of the end-to-end  TeachNet architecture, please refer to \cite{li2019vision} for more details.
Firstly, we proposed a novel criterion for generating human-robot pairings 
based on the human hand dataset Hand2.2M \cite{bighand2},
controlling the robot, and collecting joint angles and images of the 
robot in simulation.
In order to imitate the human hand pose from Hand2.2M,  we matched the Cartesian position and the link direction of the Shadow hand with the human hand pose and the proper handling of self-collisions based on  an optimized mapping method.
In the end, we collected a pairwise human-robot hand dataset, which 
includes 400K pairs of depth images,
as well as corresponding joint angles of the robot hand.

Unlike common data-driven vision-based methods which get the 3D 
human hand pose first and then map the locations to the robot,
TeachNet directly predicts robot joint angles from depth images of a human hand.
This end-to-end fashion gets rid of the time cost of post-processing but 
brings the regression challenge due to the different domains of the
human hand and the robot hand.
TeachNet tackles this challenge by using a two-branch (robot branch and 
human branch) structure and by introducing a consistency loss.
The robot branch in TeachNet plays the role of  teacher and the human 
branch that of student because the mapping from the robot
image to the joint angles is more natural as it is exactly a 
well-defined hand pose estimation problem.
At training time, we feed the pairwise human-robot images to the 
corresponding branch and the consistency loss encourages the
hand feature in the human branch to be as consistent as the hand feature in 
the robot branch. Once the TeachNet is learned, the mapping function $ f_{m} $ is thereby determined. At inference time, only the human branch is needed, accordingly, TeachNet takes an image of a human hand as input and then outputs the estimated joint angles $ q_{r} $ of the robot hand.

\subsection{Prediction of the desired force commands based on the biomimetic control strategy}

We consider that the robot hand is controlled  under the force (i.e., torque) control mode in the joint space. According to the  biomimetic control strategy in \cite{yang2011human}, the control input $ \tau_{c} $ is split into two parts including a feedforward force $ v $ and an impedance term $  u $. Therefore, we have
%\footnote{We assume the dynamics model of the robot hand is already known.}

\begin{equation}
\tau_{c} =  -u - v.
\label{controlIUPUT}
\end{equation}

The impedance term is given by
\begin{equation}
u = K_{s}e + K_{d}\dot{e},
\label{commandU}
\end{equation}
with
\begin{equation}
  \left\{
   \begin{array}{c}
   e = q -q_{d} \\
   \dot{e}= \dot{q} - \dot{q}_{d}  \\
   \end{array},
  \right.
  \end{equation}
where $ K_{s}\in R^{N_{r}\times N_{r}} $ and $ K_{d}\in R^{N_{r}\times N_{r}} $ respectively represent the stiffness and damping matrix. $ e\in R^{N_{r}\times 1} $ and $ \dot{e}\in R^{N_{r}\times 1} $ are joint angle and velocity errors. $ q_{d}\in R^{N_{r}\times 1} $ and $ \dot{q}_{d}\in R^{N_{r}\times 1} $ represent the desired joint  angle and desired velocity\footnote{In this work, the desired velocity value  of each DOF, $ \dot{q}_{d, i} $ is set to zero.}, respectively. $ q \in R^{N_{r}\times 1} $ and $ \dot{q} \in R^{N_{r}\times 1} $ are the corresponding current states.

\begin{algorithm}[t!] %算法开始 \
\caption{Online force control command generation based on the image input via teleoperation} %算法的题目 
\label{alg1} %算法的标签 
%\begin{algorithm}[1] %此处的[1]控制一下算法中的每句前面都有标号 
\KwIn{ \\
The learned optimal TeachNet model $ f_{m} $;  \\
The constant coefficients: $ Q_{k} $, $ Q_{d} $, $ Q_{v}$, and $\pi $;\\
The Gaussian basis vector $ g $. \\

}
%\KwOut { \\
%The  parametric vectors/matrices: $ \theta_{K_{S}} $, $ \theta_{K_{D}} $, $ \theta_{F} $.}
%\begin{algorithmic}
\Begin
{Initialize the parameters  $ {\theta}_{k}$, $ {\theta}_{d} $,  and $ {\theta}_{v} $; \\
\While{task not done}
{
Sense an image of the human hand pose $ I_{h} $;\\
Calculate the desired robot hand pose through Eq. \re{imageTOjoint};\\
Get the robot current states; \\
Calculate the sliding error $ \varepsilon $; \\
Update $ {\theta}_{k}$, $ {\theta}_{d} $ and $ {\theta}_{v} $ using Eq. \re{updating};\\
Generate the desired force control command $ \tau_{c} $\; 
Send the force command to the robot.
}
}
%\end{algorithmic}
\end{algorithm}

In this work, we propose to represent all the compliant profiles including the feedforward, stiffness and damping terms in the parametric space. Each of them can be individually coded as the inner product of a set of basis functions and corresponding parameters. Namely, we have
\begin{equation}
K_{s} = diag\lbrace {\theta}_{k}^{T}g \rbrace, \ \ K_{d} = diag\lbrace {\theta}_{d}^{T}g\rbrace, \ \  v = {\theta}_{v}^{T}g, 
\end{equation}
where $ {\theta}_{k} \in R^{N_{r}\times N}$, $ {\theta}_{d} \in R^{N_{r}\times N}$, $ {\theta}_{v} \in R^{N_{r}\times N}$ are the parameters corresponding to each of the compliant profiles mentioned above. And $ g \in R^{N \times 1}$ is a Gaussian basis vector and  each of its element is calculated by
\begin{equation}
[{g}]_{n} = \frac{\omega_{n}(s)}{\sum_{n=1}^{N}\omega_{n}(s)},
\end{equation}
with
\begin{equation}
\omega_{n}(s)= \exp(-0.5h_{n}(s-c_{n})^{2}),
\label{gaussian}
\end{equation}
where $ s $ is a phase variable determined by $ \dot{s} = -s $. $ c_{n} $ and $ h_{n} $ are the centers and widths of the basis, and $ N $ is the total number of the Gaussian models which is manually set in advance. 

To predict the desired control force at each time step, we need to adapt the parameters  $ {\theta}_{k}$, $ {\theta}_{d} $ and $ {\theta}_{v} $ in an online manner, based on the robot current states. To this end, we derive the updating laws according to the following cost functions.

%motivated by the principles in human motor learning control \cite{yang2011human, li2018force}. 

Firstly,  the widely used cost function in robotic control is used to minimize  the motion tracking error as follows
\begin{equation}
J_{e} = \frac{1}{2}\varepsilon^{T}M(q)\varepsilon,
\end{equation}
where the sliding error is calculated by $ \varepsilon = \dot{e} + \pi e $ with a positive constant coefficient $ \pi $, and  $M(q)\in R^{N_{r}\times N_{r}}$ is the inertia matrix.

\begin{figure}[t!]
\centering
\includegraphics[scale=0.7]{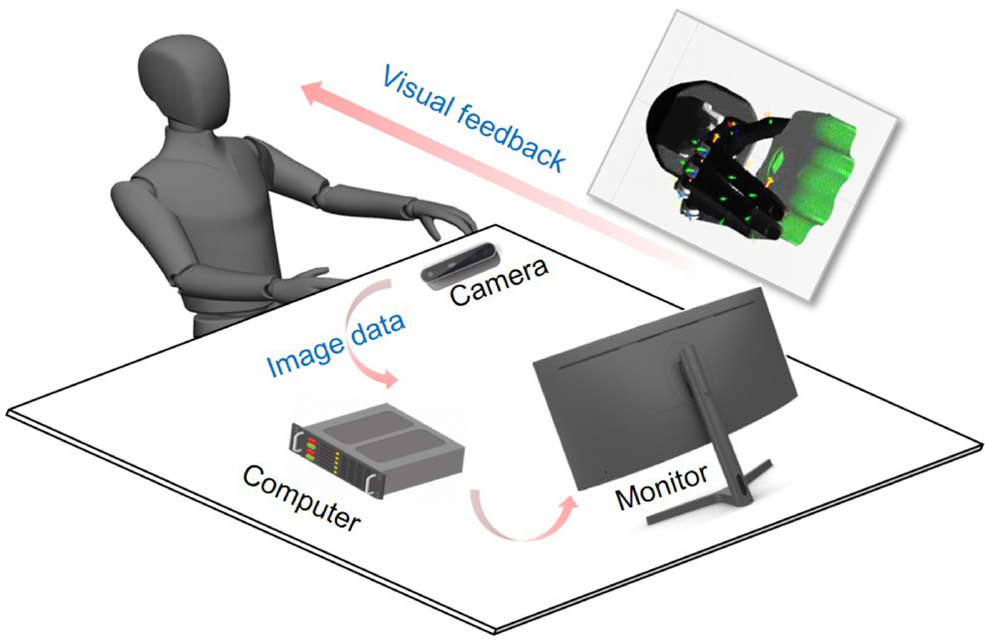}
\caption{\small The schematic illustration of the  real-sim experimental setup. }
\label{setup}
\end{figure}

Based on the principles in human motor learning control \cite{yang2011human, li2018force, zeng2021approach},  the following cost is then considered to enable the convergence to the desired parameters, i.e., $ {\theta}_{k}^{*}(t) $, $ {\theta}_{d}^{*}(t) $,  and $ {\theta}_{v}^{*}(t) $ that model the interaction dynamics,
\begin{equation}
J_{c} = \frac{1}{2}\tilde{\Phi}^{T}Q^{-1}\tilde{\Phi},
\end{equation}
where 
\begin{equation}
\begin{split}
\tilde{\Phi} = \Phi -\Phi^{*} = [\tilde{{\theta}}_{k}^{T}, 
 {\tilde{\theta}}_{d}^{T},  
 {\tilde{\theta}}_{v}^{T} ]^{T},
\end{split}
\end{equation}
with
\begin{equation}
\Phi =[{\bar{\theta}}_{k}^{T}, 
 {\bar{\theta}}_{d}^{T},  
 {\bar{\theta}}_{v}^{T} ]^{T}, 
\end{equation}
and 
\begin{equation}
\begin{split}
\Phi^{*}(t) =&[{\bar{\theta}}_{k}^{*T}, 
 {\bar{\theta}}_{d}^{*T},  
 {\bar{\theta}}_{v}^{*T}]^{T},
\end{split}
\end{equation}
%\begin{equation}
%\begin{split}
%\bar{\theta}_{k}^{*} = \frac{1}{N}\sum_{i=1}^{N}{\theta}_{k}^{*}, 
%\bar{\theta}_{d}^{*} = \frac{1}{N}\sum_{i=1}^{N}{\theta}_{d}^{*}, 
%\bar{\theta}_{v}^{*} = \frac{1}{N}\sum_{i=1}^{N}{\theta}_{v}^{*}, 
%\end{split}
%\end{equation}
where ($ \bar{.} $) denotes the row average vectors of the corresponding parameters.  The  matrix $Q $ is defined by
\begin{equation}
Q = {\rm diag}(Q_{k}\otimes {\rm I}, Q_{d}\otimes {\rm I}, Q_{v}\otimes {\rm I}), 
\end{equation}
where $ Q_{k}\in R^{N_{r}\times N_{r}} $, $ Q_{d}\in R^{N_{r}\times N_{r}} $ and $ Q_{v}\in R^{N_{r}\times N_{r}} $ are symmetric positive-definite matrices,

The updating goal is therefore to minimize the overall cost, i.e., $\min\Vert J_{c} + J_{e} \Vert  $. Finally, a standard derivation procedure\footnote{Let the derivative of the cost equal to zero to obtain the updating laws $\dot{ \theta}_{k} $, $\dot{ \theta}_{d} $, and $\dot{ \theta}_{v} $. For simplicity, we omit the detail in this paper.} is  employed to obtain the following updating laws for the $ n_{r}$-th ($ n_{r} \in [0, \cdots , N_{r}]  $) DOF, 
\begin{equation}
\begin{split}
\dot{ \theta}_{k, n_{r}}^{T} &=  Q_{k, n_{r}}\varepsilon_{n_{r}}e_{n_{r}}g \\
\dot{ \theta}_{d, n_{r}}^{T}  &= Q_{d, n_{r}}\varepsilon_{n_{r}}\dot{e}_{n_{r}}g \\  
  \dot{ \theta}_{v, n_{r}}^{T} &=  Q_{v, n_{r}}\varepsilon_{n_{r}}g.
\end{split}
\label{updating}
\end{equation}

The procedure of the generation of force control commands with the proposed approach is summarized in Algorithm \ref{alg1}.

\section{Simulation experiments}
\label{experiments}

\begin{figure}[t!]
\centering
\includegraphics[scale=0.8]{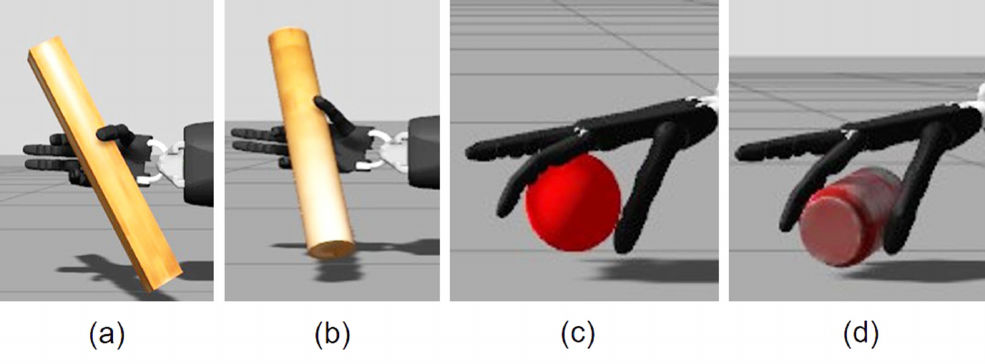}
\caption{\small The final grasping configurations for the (a) cube wood, (b) cylinder wood, (c) ball, and (d) can, under the adaptive fore mode. Three fingers (TH, FF and LF) are used in the grasping task.}
\label{finalCongfigWood}
\end{figure}
\begin{figure}[t!]
\centering
\includegraphics[scale=0.85]{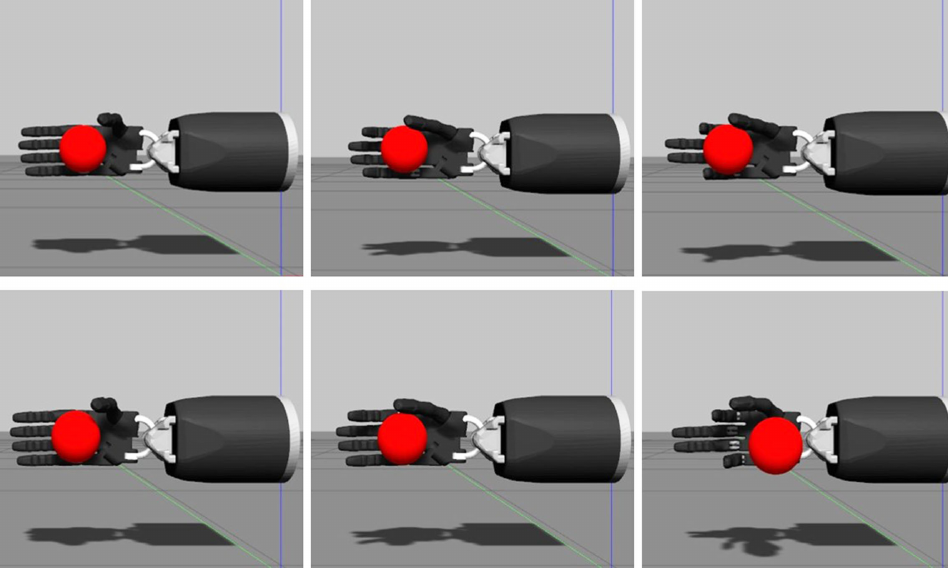}
\caption{\small The screenshots of the grasping ball task with (upper row) and without (lower row) adaptive force control. Left to Right: initial configuration, the TH contacts the ball, and final configuration.}
\label{graspBall}
\end{figure}

\subsection{Real-Sim simulation setup}
A real-sim setup is established for the simulation experiments (see Fig.~\ref{setup}). During each task, the human demonstrator adjusts the hand pose to guide the simulated robot hand to complete the task under the visual feedback. A camera (Intel RealSense F300) is used to capture the depth images of the human hand, based on which the desired robot hand pose can be estimated via the TeachNet model\footnote{\url{https://github.com/TAMS-Group/TeachNet_Teleoperation}}. The virtual Shadow Motor Hand in the Gazebo simulator with the ODE engine is utilized for our experiments,  based on the ROS package\footnote{\url{https://github.com/shadow-robot/sr_core}} provided by the Shadow Robot Company. This robot hand is equipped with five fingers, i.e, the thumb (TH), the first finger (FF), the middle finger (MF), the ring finger (RF), and the little finger (LF). TH and LF have five DOFs each, and the other three fingers have four DOFs each. Moreover, the wrist joint has another two DOFs. In our usage, the Shadow robot hand is torque-controlled under the TEACH mode. The simulation environment is run on the Ubuntu 18.04 system with a CPU Intel Core i5-8500 and a NVIDIA 1050 Ti GPU. The average updating time at each time step is 0.036s.

\begin{figure}[t!]
\centering
\includegraphics[scale=0.8]{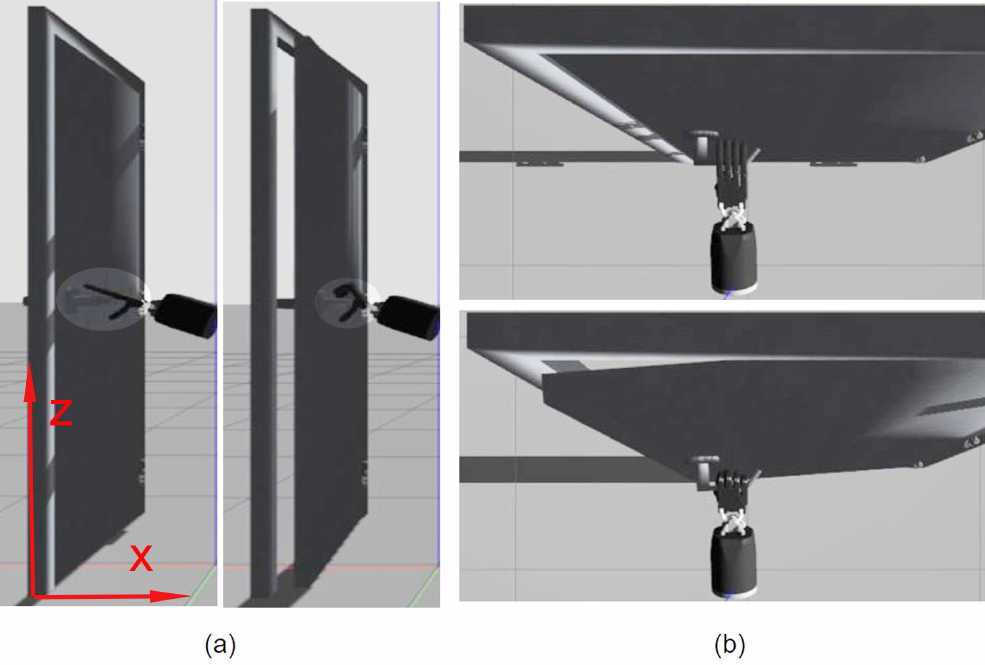}
\caption{\small The initial and final configurations during the opening-door task from the (a) side and (b) top view, respectively.}
\label{openDoor}
\end{figure}

\begin{figure}[t!]
\centering
\includegraphics[scale=0.58]{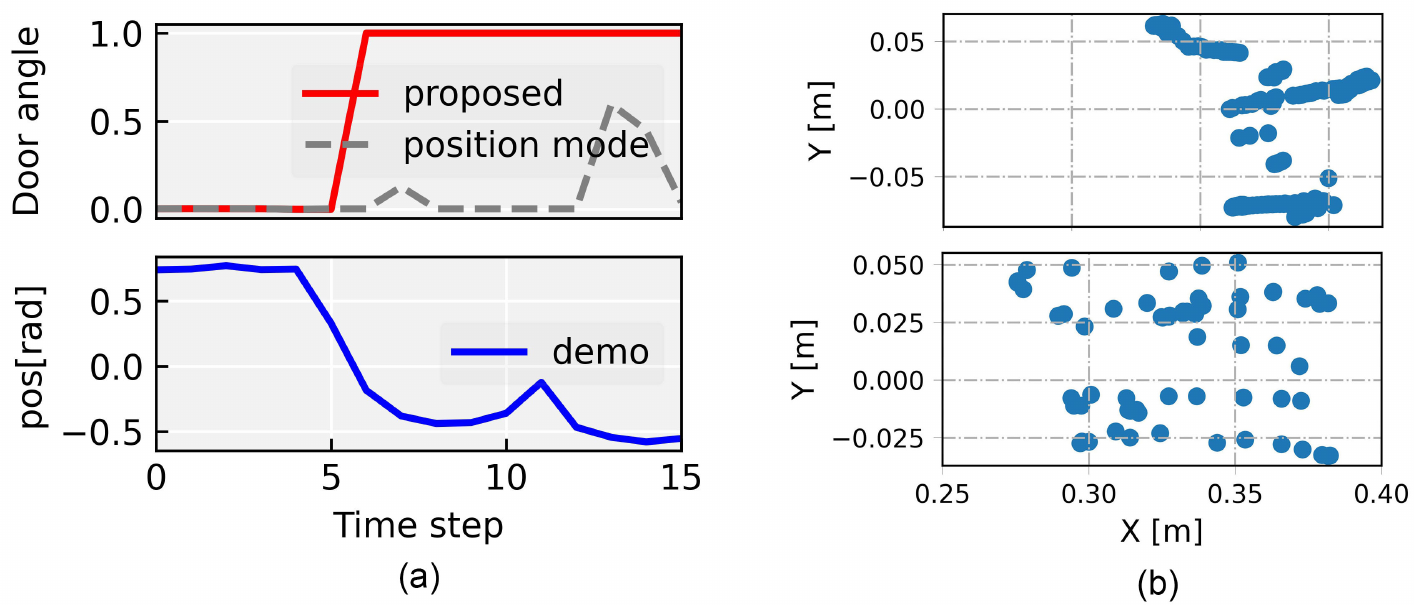}
\caption{\small (a) The upper row shows the normalized angle changes of the door with respect to the word frame, the lower row represents the general trend of the changes of the demonstrated joint angles from TeachNet. For the sake of better visualization, the door angles are normalized to $ [0, 1] $, and joint angles of the robot are reduced to one dimension using  PCA. (b) shows the positions of the contact points under the position (upper row) and force (lower row) control modes. }
\label{doorRes}
\end{figure}

\subsection{Experimental results}
%We carry out four tasks to validate the proposed approach, i.e., gasping, opening-a-door, turning-a-cap, and touching-a-mouse. 

To validate whether the proposed adaptive force control could yield better performances, we compare the proposed force control with position control on four tasks i.e., grasping, opening-a-door, turning-a-cap, and touching-a-mouse. The experimental video is available
at \url{https://www.youtube.com/watch?v=xL9BvPGIKxE}.

Under the position control mode, the position commands from TeachNet model are directly used to control the robot hand. Notely, while virtual environments are dominated by physics (e.g. object weights and surface frictions) the absence of the force feedback makes the tasks rather challenging \cite{garcia2020physics}, as even slight inaccuracies on joint angles from TeachNet may result in failure interactions.

%The results of these tasks are detailed as follows. 

\textit{Grasping task:} In this task, the robot hand is controlled to grasp four objects through teleoperation by a human demonstrator, i.e.,  a cube wood, a cylinder wood, a can, and a ball. Three fingers (i.e., TH, FF and LF) are used during the grasping  process. The grasping strategy is as follows: first, the demonstrator guides the thumb to contact the target object; Then, the thumb slightly adjusts the pose of the object in the hand space; finally, the other two fingers contact the object and thus  grasp the object stably. The final grasping configurations for grasping these four objects with the adaptive force control are shown in Fig. \ref{finalCongfigWood}. Conversely, the position control mode easily causes unstable behaviours after the initial contact, hence resulting in grasping failures. The unstable behaviours are mostly oscillations between the hand and the object in the simulation due to the contact force's rigidity, see Fig. \ref{graspBall} as an example.

\begin{figure}[]
\centering
\includegraphics[scale=0.8]{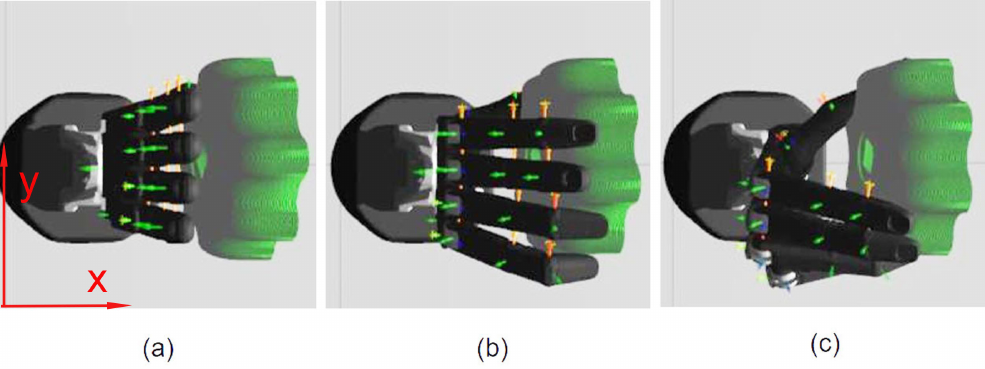}
\caption{\small The screenshots of the turning-a-cap task. (a), (b) and (c) denote the initial, middle and final configurations.}
\label{capPic}
\end{figure}

\begin{figure}[t!]
\centering
\includegraphics[scale=0.52]{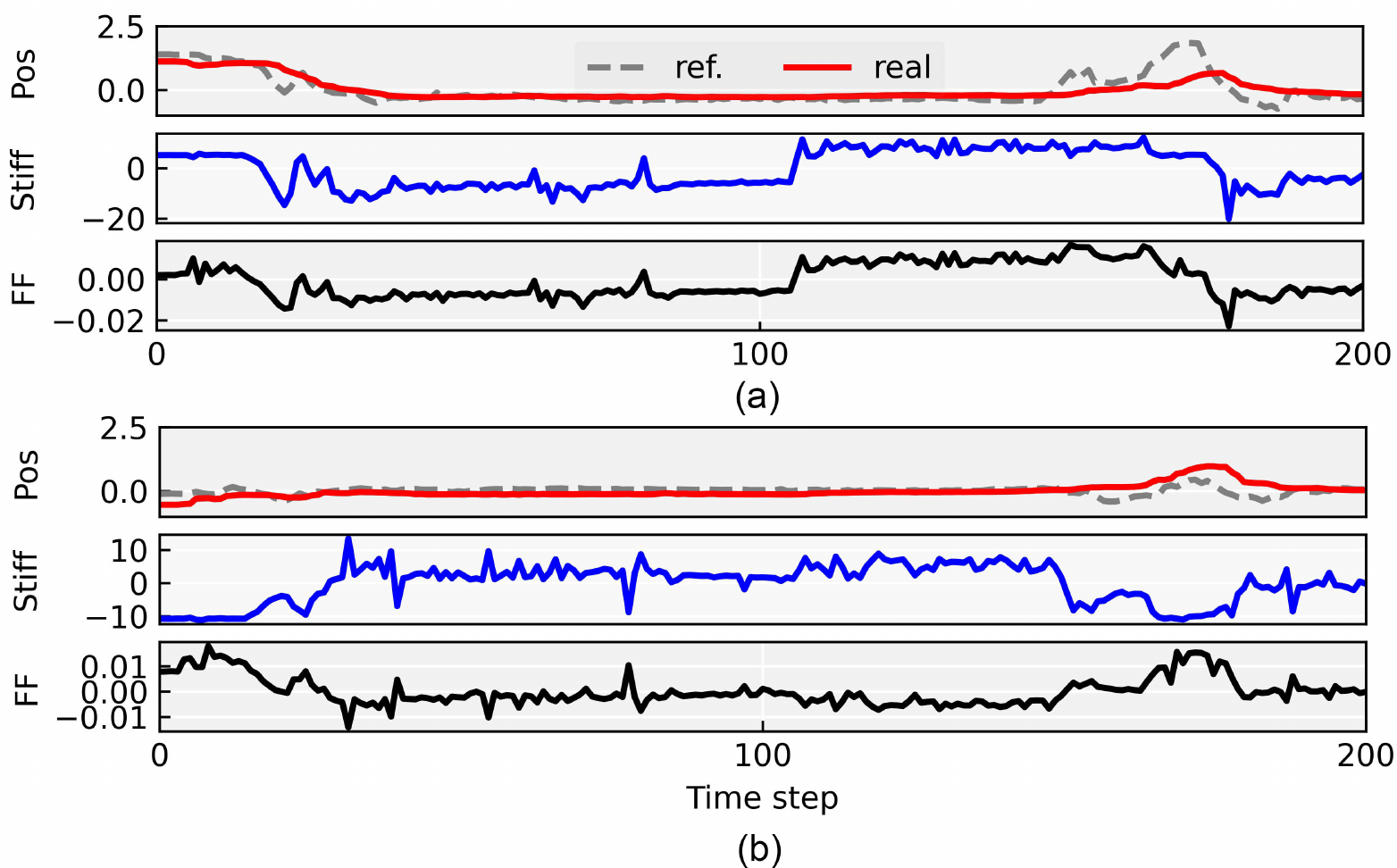}
\caption{\small The online learned compliant profiles including stiffness (stiff.) and feedforward (FF) force along the execution of the joint angles in the turning-a-cap task. All the profiles are reduced to the 2D space using PCA, the first and second components shown in (a) and (b), respectively. The reference and real curves mean the joint angles estimated from the TeachNet and collected from the robot hand, respectively.}
\label{capLines}
\end{figure}

\begin{figure}[]
\centering
\includegraphics[scale=0.5]{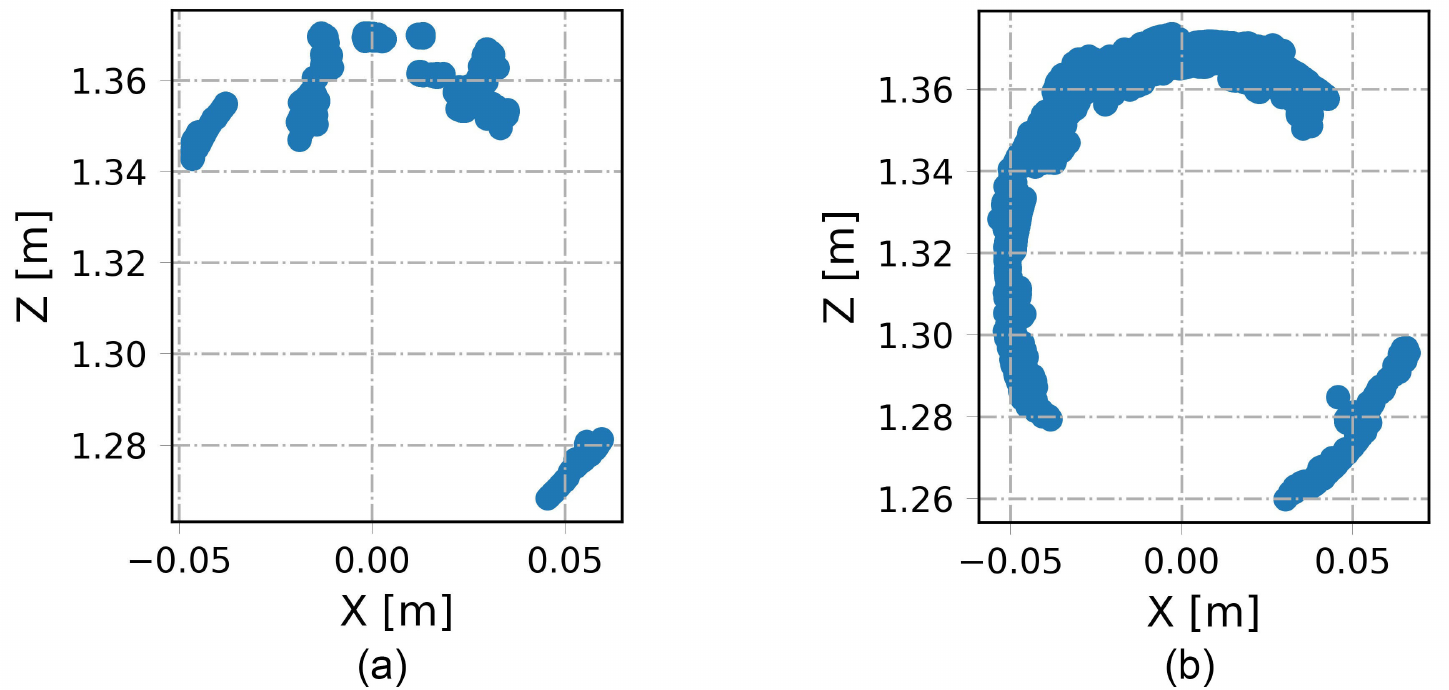}
\caption{\small The positions of the contact points in the $x-z$ plane during the turning process, under the (a) position control and (b) adaptive force control modes, respectively.}
\label{capContacts}
\end{figure}

\textit{Opening-a-door:} The second task is that the human demonstrator guides the robot hand to open a door by pulling the handle in the $x$ direction. All these five fingers are used in this task. Since the base and the wrist of the Shadow hand are fixed, we can only rely on the fingers' interaction with the handle to open the door. We find that the robot can open the door smoothly, with the proposed adaptive force control strategy, as depicted in Figs. \ref{openDoor} and  \ref{doorRes}. On the other hand, under the position mode, the teleoperator only can occasionally open the door and fails to make the door open as wide as under the force control mode. Fig. \ref{doorRes}(b) shows the positions of the contacts between the robot hand and the door handle in the $ x-y $ plane. Under the adaptive force control  mode, we can see the contact points are almost evenly distributed along the $ x $ axis, which suggests the stable interaction between the robot hand and the handle during the execution of the task. This maybe because our control strategy can well deal with the interaction dynamics of the Gazebo simulator.

\textit{Turning-a-cap:} In this task, the  robot hand is teleoperated by the human demonstrator to turn a cap using five fingers. The frame of the cap is fixed in the Gazebo world, and the cap can be rotated in the $x-z$ plane. In a real-world task like turning a cap, humans need to adapt the motion of both arm and hand coordinately to  complete this task. More importantly, the rotation of the wrist joint plays a key role during the turning process. In our teleoperation system, however, the fixed base and wrist of the Shadow hand make this task more challenging than usual. 
%the robot hand has to be fixed in the world  frame and the tracking of the human demonstrator's hand wrist motion can not be well achieved via the TeachNet model. These two limitations make this task more challenging than as usual. Therefore, the success of the task has to highly depend on the robot hand's dexterous manipulation ability. 
%We also compare the proposed force control strategy with the direct motion control.
The robot hand is guided to make contact with the cap using a proper configuration and then to adapt the movements of all the fingers to turn the cap. We observe that the fingers can move coordinately and cooperate well with each other to complete the task using the proposed force control strategy  (see Fig. \ref{capPic} as an example). Then we analyse the compliant profiles including the stiffness and feedforward force learned online with the execution of the task. For better illustration, we reduce the stiffness and the feedforward force of all joints to the 2D space using the PCA algorithm. The results visualized in Fig. \ref{capLines} indicate that the robot hand (the reference curves) can track the human hand (the real curves) with the online adaptation of the stiffness and feedforward force profiles based on the pose difference between the human and the robot hands. Under the direct position control mode, however, the robot hand fails to turn the cap due to the lack of coordination and dexterity. Furthermore, the distribution of the contact points  obtained under the adaptive force mode is more caplike (see  Fig. \ref{capContacts}).

 %The compliant profiles including the stiffness and feedforward force, learned online with the execution of the robot hand joint angles, are reduced to the 2D space using the PCA algorithm and they are visualized in Fig. \ref{capLines}. It indicates that our approach can enable the robot hand to track the human pose, with the online adaptation of the compliant profiles. Under the direct position control mode, however, the robot hand fails to turn the cap due to the lack of  manipulation dexterity. Furthermore, the distribution of the contact points  obtained under the force mode is more caplike, as shown in Fig. \ref{capContacts}. 
 
 %shown in Fig. \ref{capContacts} is less similar to the caplike. 
 
% which can further confirm the different performances under these two control modes. 

\begin{figure}[t]
\centering
\includegraphics[scale=0.82]{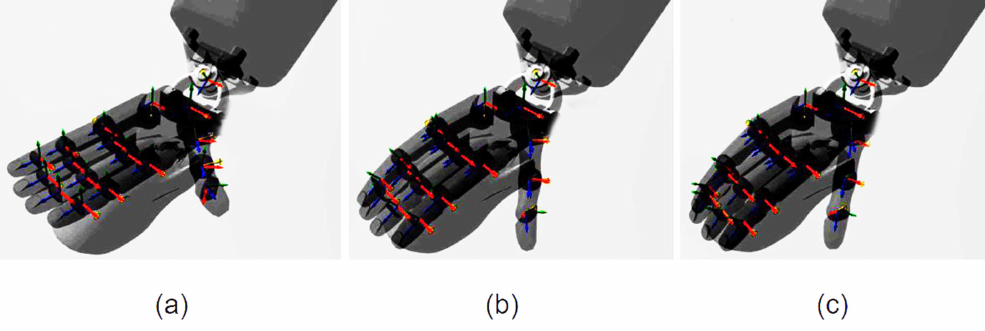}
\caption{\small The screenshots of the touching-a-mouse task. (a), (b) and (c) denote the initial, middle and final configurations.}
\label{mousePic}
\end{figure}

\textit{Touching-a-mouse:} To further explore the compliance with the adaptive force control, we investigate the performances when   the robot hand contacts with a curved surface by touching a mouse, as shown in Fig. \ref{mousePic}. We mainly focus on  achieving  stable contacts between the hand and the mouse surface  with small contact forces. Namely, we expect that the robot hand is able to touch the surface of the target object in a more human-like manner. To evaluate the impact of the adaptive control strategy, the task is conducted under three different control strategies: (a) with the proposed adaptive control; (b) with force control but with a fixed-gain-based impedance controller which has been often applied in robotics;  and (c) the position control mode. Under each condition, the task is repeated  ten times. During each test, the contact points and forces are recorded for evaluation of the performances. 

\begin{figure*}[t!]
\centering                                                          
\subfigure[]{                   
%\begin{minipage}{7cm}
\centering                                                       
\includegraphics[scale=0.53]{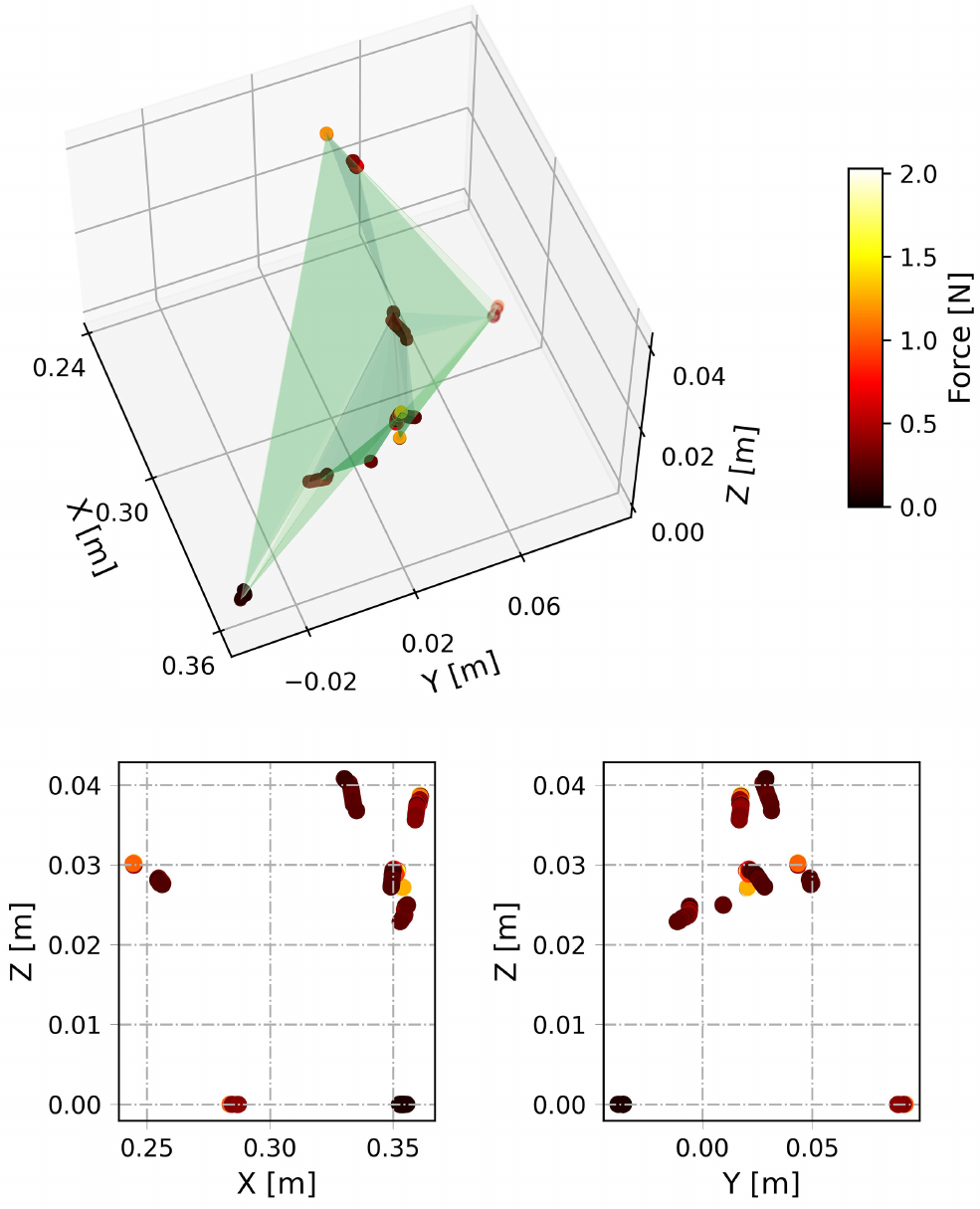}}            
%\end{minipage}}
\subfigure[]{                    
%\begin{minipage}{5cm}
\centering                                                          
\includegraphics[scale=0.53]{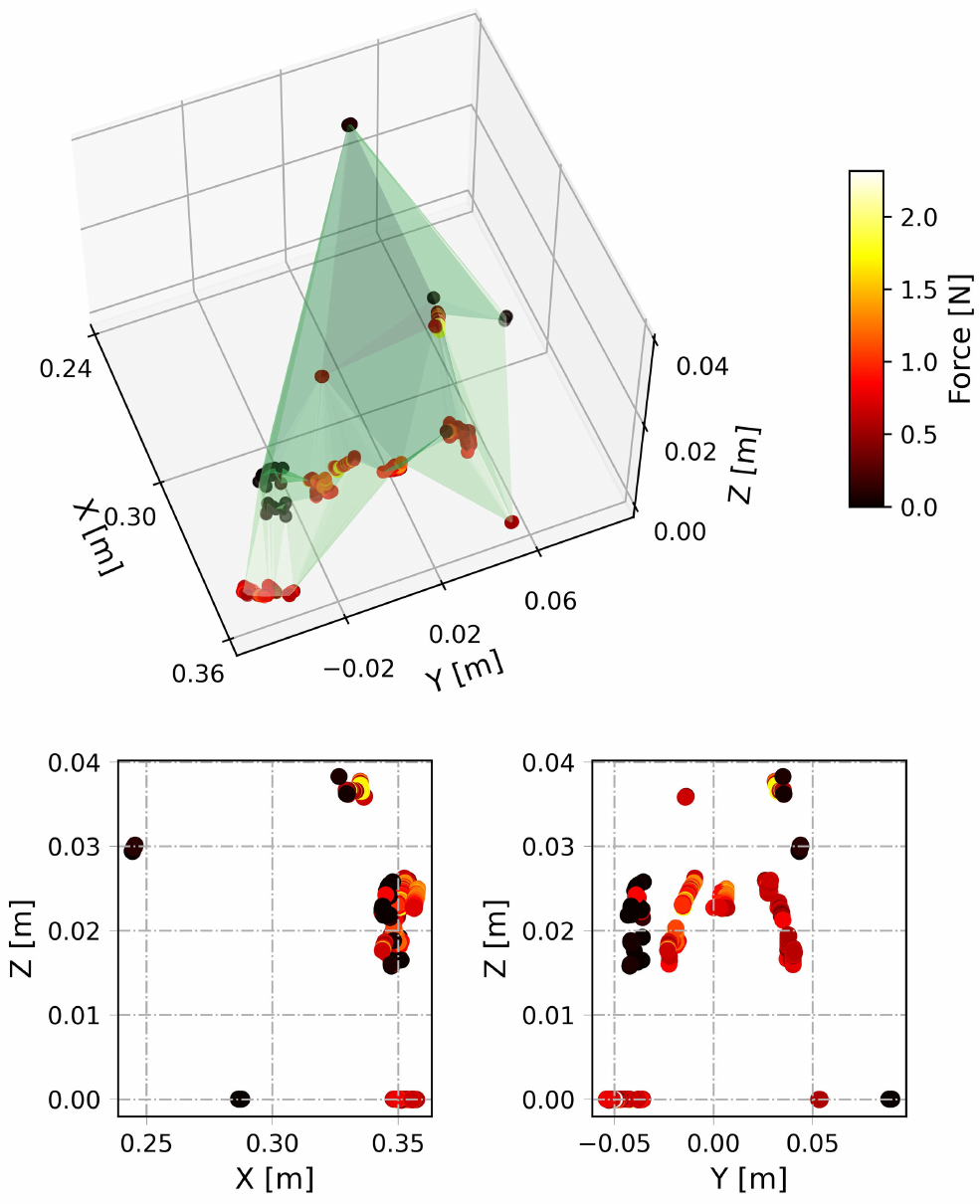}}              
%\end{minipage}}
\subfigure[]{                    
%\begin{minipage}{5cm}
\centering                                                          
\includegraphics[scale=0.38]{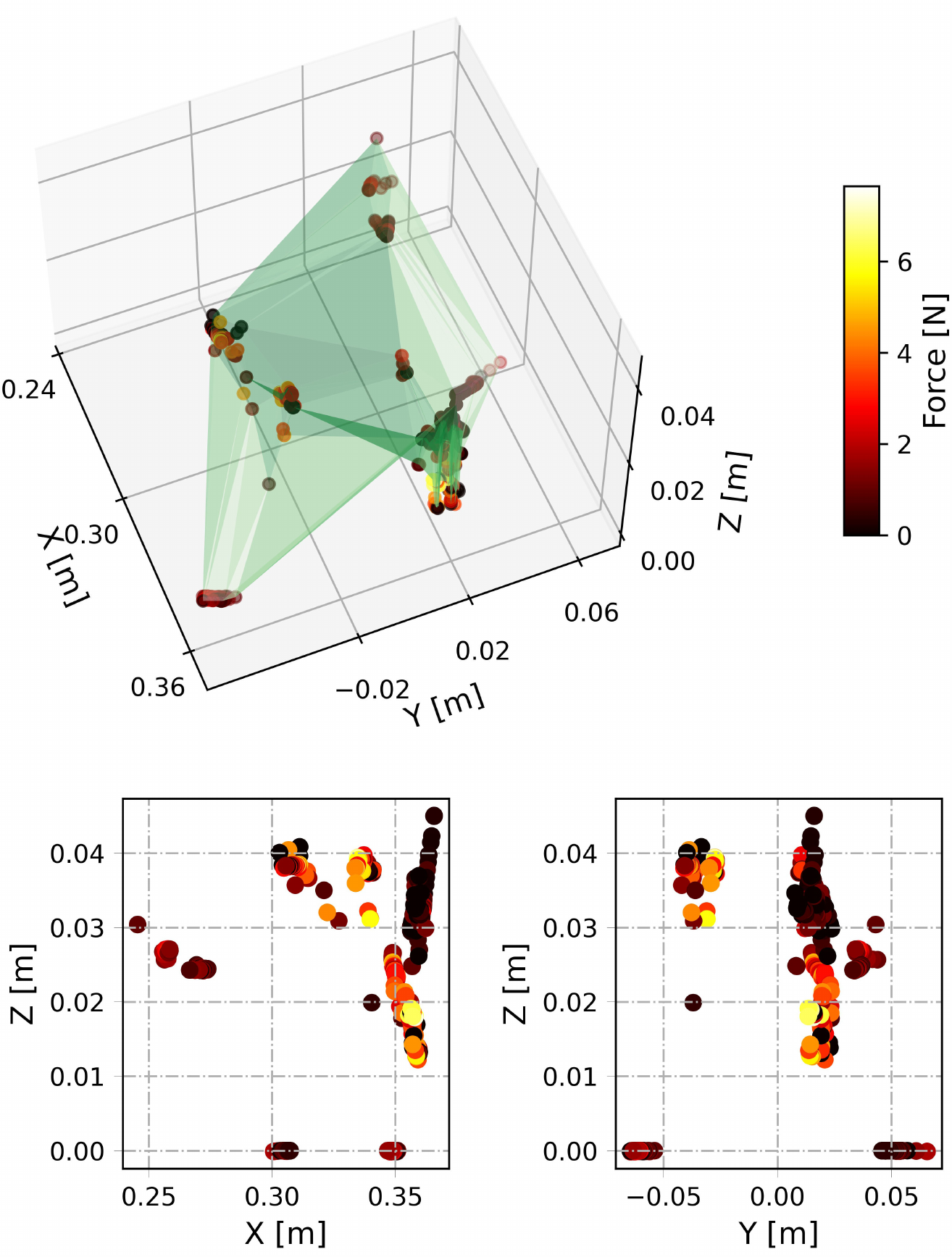}}              
%\end{minipage}}
\caption{\small The distribution of the contact points and the contact force  during the touching process under the (a) adaptive force control, (b) fixed-gain based force control, and (c) position control conditions, respectively. The upper row show the results in the 3D space, and the lower row shows the corresponding results which are projected to the $ x-z $ and $ y-z $ planes. }                     
\label{mouseContact}                                                      
\end{figure*}

\begin{figure}[t]
\centering
\includegraphics[scale=0.65]{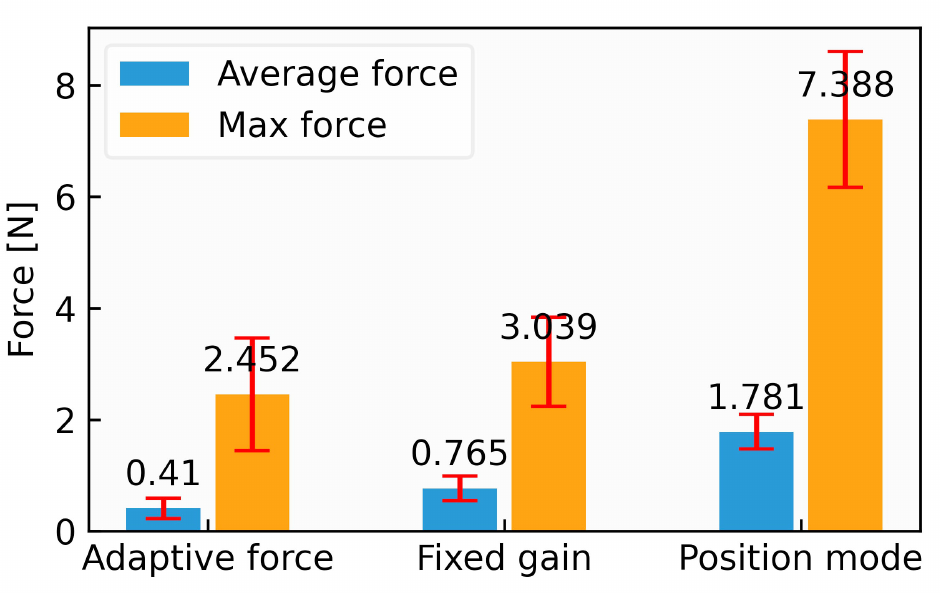}
\caption{\small The max and average contact forces during the touching phase under the three control conditions.}
\label{mouseForce}
\end{figure}

Fig. \ref{mouseContact} manifests that under the control condition (a) the contact points of each local region are distributed in a more clustered way than that under the other two control conditions, with comparatively low contact forces. Under condition (c), there are obvious  slippery points  with larger contact forces, due to the rigid interaction with the mouse of the robot hand. Condition (b) obtains a moderate performance with several slippery contact points, although the contact forces are smaller than the case under the direct position control mode. We collect the contact forces from these tests under each control condition, and calculate the maximum and average forces. The results (Fig.~\ref{mouseForce}) demonstrate significantly lower average as well as maximum forces with our proposed control strategy.

%as shown in Fig. \ref{mouseForce}.  The maximum force obtained under the condition (a) is $30 \%$ and $80 \%$ of that obtained under conditions (b) and (c), respectively, and the average force is $23 \%$ and $54 \%$ of that obtained under the control conditions (b) and (c), respectively. 

%\begin{figure}[t!]
%\centering                                                          
%\subfigure[]{                   
%%\begin{minipage}{7cm}
%\centering                                                       \includegraphics[scale=0.07]{pic/mouseRes0.jpg}}            
%%\end{minipage}}
%\subfigure[]{                    
%%\begin{minipage}{5cm}
%\centering                                                          
%\includegraphics[scale=0.07]{pic/mouseRes1.jpg}}              
%%\end{minipage}}
%\subfigure[]{                    
%%\begin{minipage}{5cm}
%\centering                                                          
%\includegraphics[scale=0.07]{pic/mouseRes2.jpg}}              
%%\end{minipage}}
%\caption{}                     
%\label{cuttingRES}                                                       
%\end{figure} 

\section{Conclusion and Future Work}
 
 %we seek to go beyond \cite{yang2011human,li2018force} to learn compliant skills not directly in the trajectory level but in the parametric space (see Sec. \ref{LearningofHlCMPs} ). 
This work proposes a novel approach for robotic compliant grasping and manipulation based on an adaptive force control strategy through teleoperation.  Our approach takes a depth image of the human hand  as the input and predicts the desired force control commands, instead of outputting the  motion control policies directly. To enable grasping and manipulation in a  dexterous manner, the robot hand is controlled under the force/torque control mode. The proposed strategy adapts the compliant profiles (impedance and feedforward) in the force controller online,  based on the pose difference between the human hand and the robot hand step-by-step. Four types of tasks in the simulation environment Gazebo (i.e., gasping, opening-a-door, turning-a-cap, and touching-a-mouse) have been conducted to verify the effectiveness of our approach. The results show that it achieves  better performances than the position control  and the fixed-gain-based force control modes.

In  future work, we will implement our approach on a real Shadow motor hand that has been equipped with the interface for force/torque control in the joint space. Since the joints of the real Shadow hand are driven by tendons, the predicted torques from the controller need to be mapped properly to the strain values of the tendons. This might be  achieved by learning a  regression model to find the mapping between them. It is also worth noting that our model-free approach can be applied to other  torque-controlled robot hands. One limitation of our approach is the lack of tactile feedback. The interaction force  between the robot hand and its environment can be estimated from tactile signals collected from the tactile sensors mounted on the tips of the Shadow motor hand. The estimated force information can then be included in the control loop as a feedback variable to improve the interaction  dexterity.

\small
\footnotesize
\bibliography{ref}

\bibliographystyle{IEEEtran}

%\ifCLASSOPTIONcaptionsoff
%  \newpage
%\fi

\end{document}